\documentclass[fleqn,10pt,twocolumn]{AROB-ISBC-SWARM24}
\usepackage{cite}
\usepackage{amsmath,amssymb,amsfonts}
\usepackage[noend]{algorithmic}
\usepackage{algorithm}
\usepackage{graphicx}
\usepackage{textcomp}
\usepackage{xcolor}
\usepackage{amsmath,amsfonts}
\usepackage{algorithm}
\usepackage{color}
\usepackage{array}
\usepackage[caption=false]{subfig}
\usepackage{textcomp}
\usepackage{stfloats}
\usepackage{url}
\usepackage{verbatim}
\usepackage{graphicx}
\usepackage{cite}
\usepackage{amsmath,amssymb}
\usepackage{makecell}
\usepackage{bbding}
\usepackage{multirow} 
\usepackage{multicol} 
\usepackage{arydshln}
\usepackage{booktabs}
\usepackage[hyphenbreaks]{breakurl}
\usepackage{graphicx}
\usepackage{float}
\usepackage{svg}  
\title{Neural-Network-Driven Method for Optimal Path Planning via High-Accuracy Region Prediction}

\author{Yuan Huang${}^{1}$, Cheng-Tien Tsao${}^{2}$, Tianyu Shen${}^{3\dagger}$, and Hee-Hyol Lee${}^{4}$}
\speaker{Tianyu Shen}

\affils{${}^{1234}$Graduate School of Information, Production and Systems,
	Waseda University, Kitakyushu, Japan\\
(E-mail: ${}^{1}$gakki@toki.waseda.jp, \\
${}^{2}$ct.tsao@akane.waseda.jp, \\
${}^{3}$tianyushen@akane.waseda.jp, \\
${}^{4}$hlee@waseda.jp)\\
}
\abstract{%
Sampling-based path planning algorithms suffer from heavy reliance on uniform sampling, which accounts for unreliable and time-consuming performance, especially in complex environments. Recently, neural-network-driven methods predict regions as sampling domains to realize a non-uniform sampling and reduce calculation time. However, the accuracy of region prediction hinders further improvement. We propose a sampling-based algorithm, abbreviated to Region Prediction Neural Network RRT* (RPNN-RRT*), to rapidly obtain the optimal path based on a high-accuracy region prediction. First, we implement a region prediction neural network (RPNN), to predict accurate regions for the RPNN-RRT*. A full-layer channel-wise attention module is employed to enhance the feature fusion in the concatenation between the encoder and decoder. Moreover, a three-level hierarchy loss is designed to learn the pixel-wise, map-wise, and patch-wise features. A dataset, named Complex Environment Motion Planning, is established to test the performance in complex environments. Ablation studies and test results show that a high accuracy of 89.13\% is achieved by the RPNN for region prediction, compared with other region prediction models. In addition, the RPNN-RRT* performs in different complex scenarios, demonstrating significant and reliable superiority in terms of the calculation time, sampling efficiency, and success rate for optimal path planning.}

\UseRawInputEncoding
\begin{document}

\maketitle


\section{Introduction}

\begin{figure}[t]
	\centering
	\subfloat[RRT* using 0.94 second.]{\includegraphics[width=3.1in]{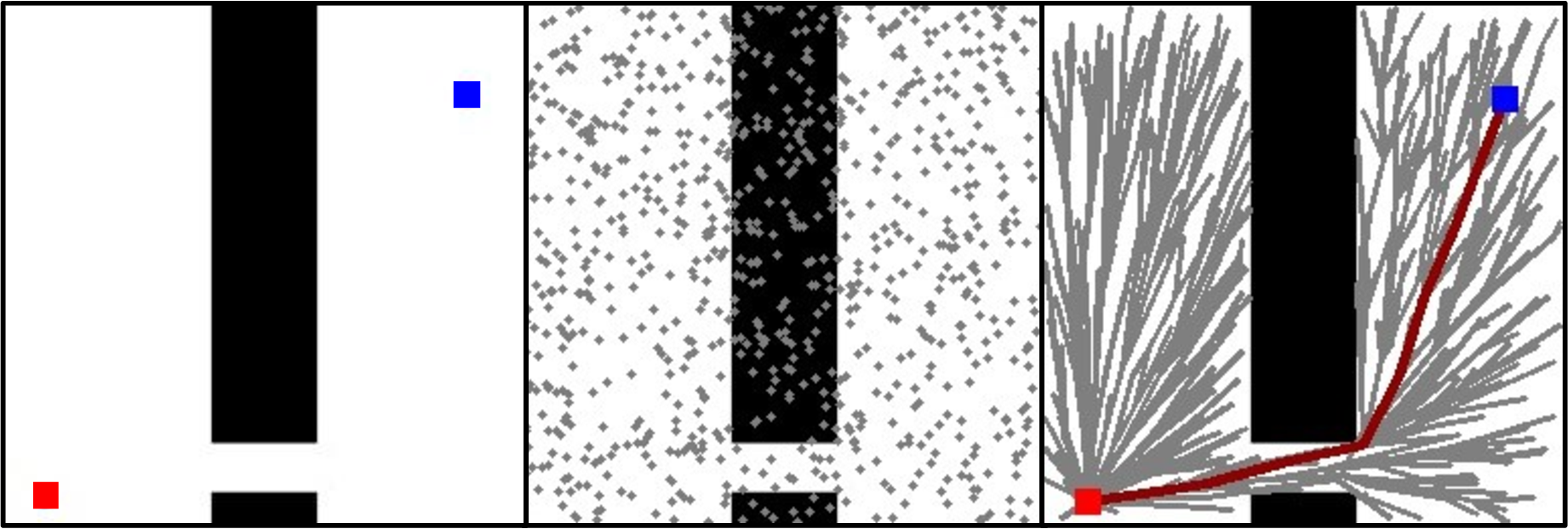}\label{fig1a}}
	\vspace{-0.3mm} 
	\subfloat[NEED-RRT* using 0.17 second.]{\includegraphics[width=3.1in]{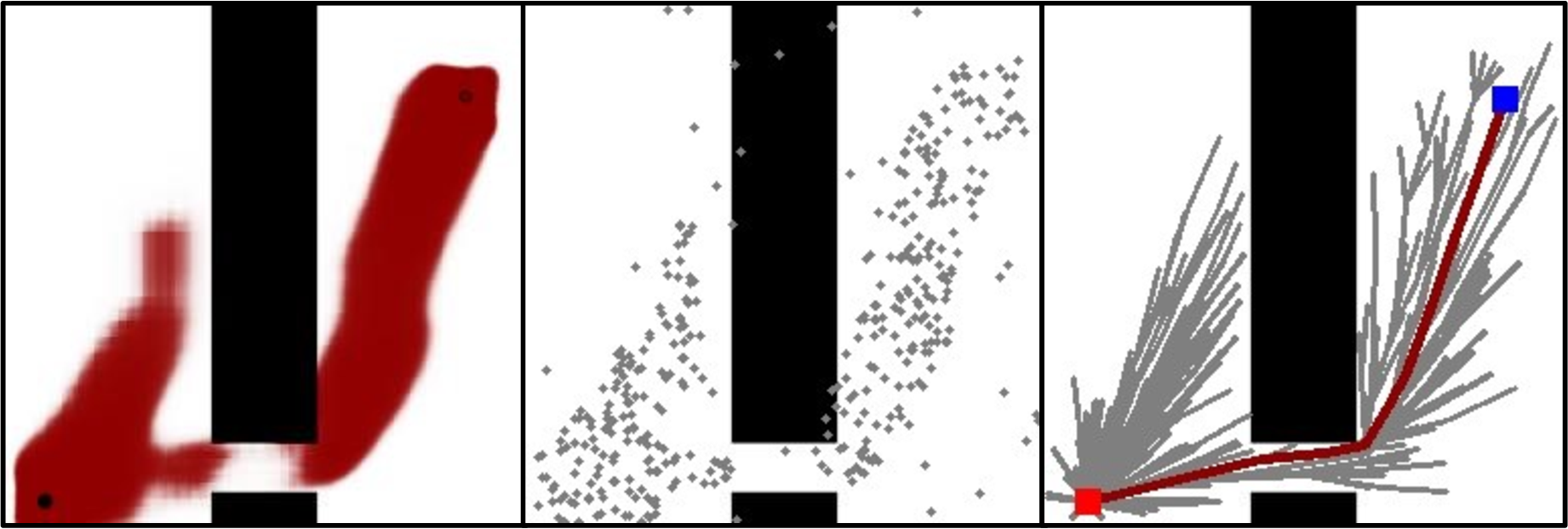}\label{fig1b}}
	\vspace{-0.3mm} 
	\subfloat[RPNN-RRT* using 0.13 second.]{\includegraphics[width=3.1in]{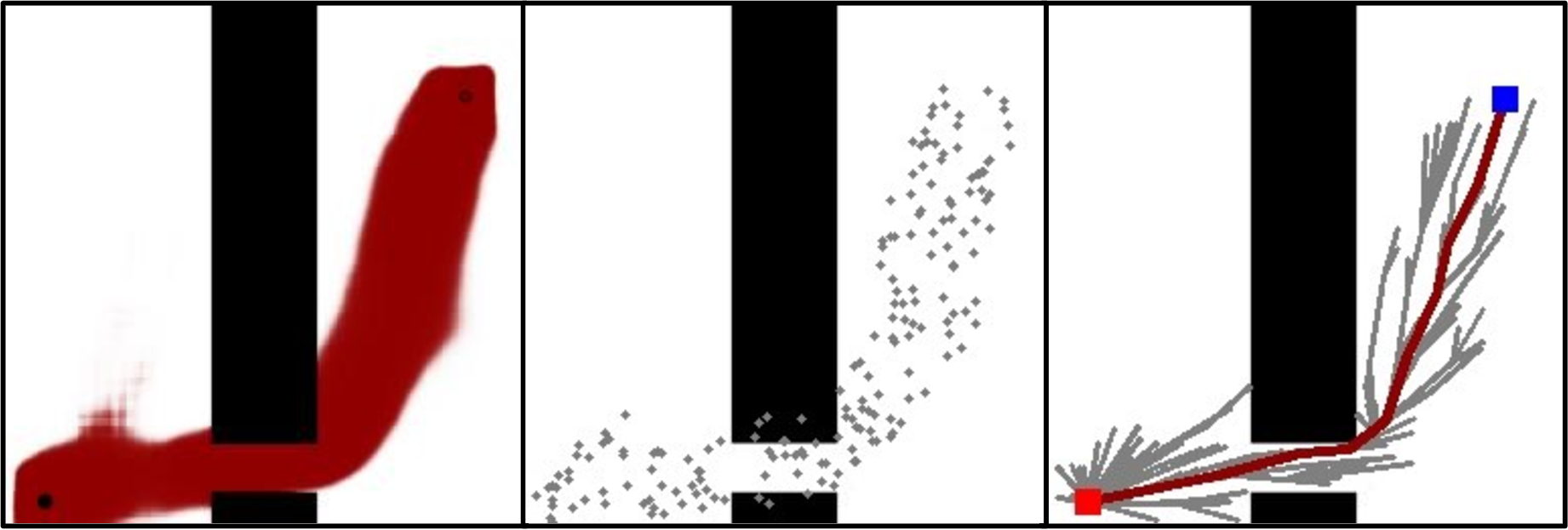}\label{fig1c}}
	\vspace{-0.3mm} 
	\caption{Sampling distributions and exploration areas of the RRT*, NEED-RRT*\cite{wang2021robot}, and RPNN-RRT*. Deep red areas represent the predicted regions based on neural-network-driven prediction (NEED)\cite{wang2021robot} and RPNN. Gray vertices are generated to find the optimal path between a start vertex (red) and a goal vertex (blue).} 
	\label{fig1}
	\vspace{-5mm} 
\end{figure} 
Optimal path planning problem involves finding a collision-free and lowest-cost path from a starting vertex to a goal vertex in a configuration space. As a fundamental problem for robots to perform various tasks, traditional path planning algorithms can be classified into two broad categories: grid-based algorithms such as Dijkstra \cite{dijkstra2022note} and A* \cite{hart1968formal}, and sampling-based algorithms like Rapidly-exploring Random Tree (RRT) \cite{LaValle1998RapidlyexploringRT} and its variant RRT* \cite{karaman2011sampling}. While the grid-based algorithms are time-consuming to implement and inappropriate for large-scale or high-dimensional problems \cite{zafar2018methodology}, the sampling-based algorithms can rapidly explore the configuration space without the requirement for discretization. As a representative sampling-based algorithm, RRT* constructs an exploration tree iteratively to find the optimal path. It guarantees asymptotic optimality and probabilistic completeness by generating numerous samples to cover the entire configuration space. However, RRT* and similar sampling-based algorithms are inefficient in complex environments due to direct reliance on uniform samples (as shown in Fig. 1(a)), which involve unnecessary exploration and lead to an increase in calculation time as the solution converges incrementally to an optimum.

To improve the sampling efficiency and reduce calculation time, several heuristic sampling methods are discussed. Urmson and Simmons \cite{urmson2003approaches} add a goal bias to generate samples, and Gammell et al. \cite{gammell2014informed} build a heuristic ellipsoidal set to shrink the whole configuration domain to subsets. A Batch Informed Trees (BIT*) \cite{gammell2020batch} method is presented by creating a series of implicit random geometric graphs as heuristics. Wang et al. \cite{wang2019finding} implement a Gaussian mixture model to provide an approximate moving point’s probabilistic heuristic, which can guide the exploration tree’s growth. 

Recently, a great deal of research has focused on utilizing neural networks to predict regions as sampling domains that surely contain the optimal path. Subsequently, neural-network-driven methods generate samples in the region to reduce the calculation time. In \cite{ichter2018learning}, a conditional variational autoencoder is designed to predict the sampling domains for non-uniform sampling. In \cite{zhang2018learning}, ambiguous sampling distributions are learned from a policy-search method to realize rejection sampling schemes for optimal path planning algorithms \cite{zhang2018learning}. Wang et al. \cite{wang2020neural} use an encoder-decoder architecture for learning from A* results, and a promising region is predicted to guide the RRT*. In \cite{zhang2021generative} and \cite{ma2021conditional}, generative adversarial network (GAN) models are presented to segment regions, where a feasible path probably exists. Liu et al. \cite{liu2021learning} present an auto-encoder-decoder-like convolutional neural network to predict regions in complex environments, and the prediction results efficiently guide the searching process. Neural-network-driven methods enable more samples to locate near the optimal path, assisting the exploration to rapidly converge to the optimum. However, one of the limitations is the unsatisfied accuracy of the prediction results, leading to problems such as dispensable samples, extensive explorations, and even failure to find a path, which is rarely discussed before. Previous methods adopt a hand-crafted heuristic-biased sampling strategy to extenuate the negative influence of inaccurate prediction results and guarantee the probabilistic completeness of algorithms. Despite that, the search process is still plagued by the low-accuracy regions with a long calculation time, especially in complex environments. Fig. 1 shows the difference in the sampling distributions and the exploration areas among RRT* and neural-network-driven methods. Misleading samples are derived from the inaccurate region in Fig. 1(b), causing an excessive and meaningless exploration with a long calculation time.

In this work, we aim to improve the accuracy of the predicted region for neural-network-driven methods, further enhance the sampling efficiency and reduce the calculation time for the optimal path planning. Our main contributions are three-fold:
\begin{itemize}
	\item{Devising a new Region Prediction Neural Network (RPNN) to generate high-accuracy regions. A full-layer channel-wise attention module is equipped with the network to extenuate the segmentation gap. A patch-level loss, denoted as purity loss, is designed to encourage the network to concentrate on the connectivity at a patch level. We evaluate the accuracy performance of the RPNN based on 3200 complex scenarios. Results show that the RPNN achieves 89.13\% accuracy in predicting the region, which outperforms other region prediction models.}
	\item{Designing a neural-network-driven optimal path planning algorithm, abbreviated to RPNN-RRT*. The high-accuracy regions guarantee outstanding performance in complex scenarios referring to the calculation time. Simulation results verify that the RPNN-RRT* reduces 29-73\% and 90-91\% calculation time compared with other neural-network-driven methods and RRT*, respectively. }
	\item{Establishing a dataset, Complex Environment Motion Planning (CEMP), via a novel data generation strategy, which contains complex scenarios with narrow passages to train and test the performance of the RPNN and other region prediction models.}\\
\end{itemize}

The rest of the paper is organized as follows. In Section 2, basic knowledge of the RRT* and limitations of previous region prediction models are introduced. A structure of the RPNN and an outline of the RPNN-RRT* are presented in Section 3. Then, we introduce the dataset, CEMP, and the data generation strategy. Moreover, experiments are conducted to assess performance on the accuracy of the prediction results and the optimal path planning in Section 4. Finally, a conclusion of the paper is drawn in Section 5.

\section{Preliminaries}
In this section, we give a brief introduction to the RRT* in Section 2.1. Besides, we discuss the limitations of the region prediction models in Section 2.2.
\subsection{RRT* and sampling strategies}
RRT* is an asymptotically optimal variant of the RRT for the optimal path planning, which was first introduced by LaValle \cite{karaman2011sampling}. Both algorithms work by constructing a tree in the configuration space, with a start vertex as a root, and iteratively adding new vertices to the tree by connecting them to existing vertices. The critical difference between RRT and RRT* is that RRT* includes two optimization steps to improve present solutions. The cost-to-come (the total costs from the current vertex tracing back to the start vertex) determines the best parent for each new vertex and allows nearby vertices to rewire their parents based on the new vertex. As shown in Algorithm 1, an exploration tree $T={(V,E)}$ is initialized and grows via subroutines $UniformSampler()$, $Nearest()$, and $Steer()$.  If a new vertex is successfully added, the function $ChooseParent()$ performs to connect the vertex to the best parent, referring to the cost-to-come. $Rewire()$ reconstructs the connections nearby the new vertex. These optimization steps help to ensure asymptotic optimality. Concisely, it can approach the optimum as the sample number increases indefinitely.

As the uniform sampling cannot efficiently search in complex environments, non-uniform sampling strategies are designed. The motivation is that more samples should be generated in a certain region which is more important. In general, the region is produced through some rejection rules, which indicates a requirement for the uniform sampling at the initial search stage. With the development of deep learning, the region prediction is regarded as a segmentation task. Various segmentation models are implemented to predict the region based on numerous path-planning instances. The neural-network-driven methods with non-uniform sampling have revealed great improvement in respect of the reduction of the calculation time and the sampling efficiency. The proposed method RPNN-RRT* adopts the non-sampling strategy with a heuristic sampling bias $b_h$ (as shown in Algorithm 1) to explore the configuration space rapidly. Overall structures of the RRT* and the RPNN-RRT* are outlined in Algorithm 1.

\begin{figure}[t]
	\centering
	\includegraphics[width=3.3in]{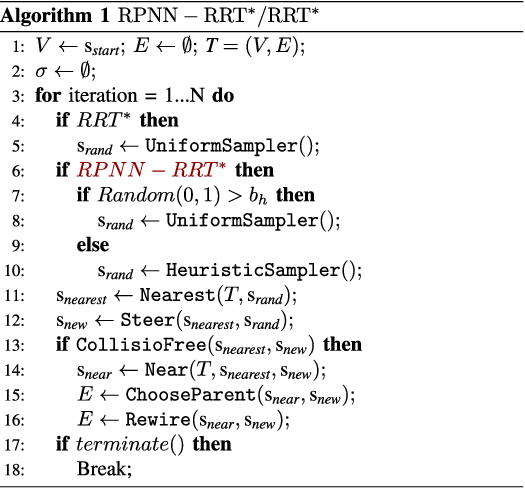}\label{alg1}
	\vspace{-1.5em}
\end{figure} 
\subsection{Limitations of region prediction models}
Recent learning-based methods \cite{liu2021learning,ma2022enhance,wang2021robot,wang2021deep} adopt a UNet-like auto-encoder-decoder architecture to predict the regions. The architecture is first discussed in \cite{cho2014learning} to segment images, and Ronneberger et al. \cite{ronneberger2015u} modify it by building a skip connection between the encoder and the decoder outputs. The skip connection builds an expansive path to aggregate context information from the encoder to the decoder directly. Rich propagated spatial information helps the decoder to recover the original resolution with high accuracy. However, Wang et al. \cite{wang2022uctransnet} point out that semantic gaps give rise to incompatible and ambiguous feature sets, which negatively influence the accuracy of the prediction. How to fuse the contextual information and narrow the semantic gaps? Features are not holding equal contributions in channel-wise, while some channels play a more crucial role in semantic fusion. Driven by the important principle, we conduct a UNet-like model with a full-layer channel-wise attention mechanism to efficiently fuse the shallow and deep feature information. 

In addition to the semantic gaps, we further investigate the other potential reasons behind the unsatisfied prediction results. Research \cite{wang2020neural,zhang2021generative,ma2021conditional,liu2021learning,wang2021robot,ma2022enhance, wang2021deep} utilize a set of suboptimal paths or only an optimal path as ground truth. However, the ground truth in datasets should have a homologous association with the prediction results, while the ground truth is expected to be explicit regions as prediction targets. Moreover, suboptimal paths as the ground truth are unavailable to quantify and evaluate the performance of prediction results \cite{wang2021robot}. We utilize a reasonable data generation strategy to build the CEMP for training, while the ground truth as well as regions precisely cover the optimal solutions. Accordingly, the performance of prediction results can be quantifiable and assessed through supervised learning.

\section{Methodology}
In this section, we first present details of the RPNN to predict promising regions, including modifications made to an encoder, a decoder, and a deep supervision module. In Section 3.2, we introduce the novel loss function to improve the accuracy performance of the prediction results. Finally, we outline the procedure of the RPNN for optimal path planning.
\begin{figure}[t!]
	\centering
	\includegraphics[width=3.3  in]{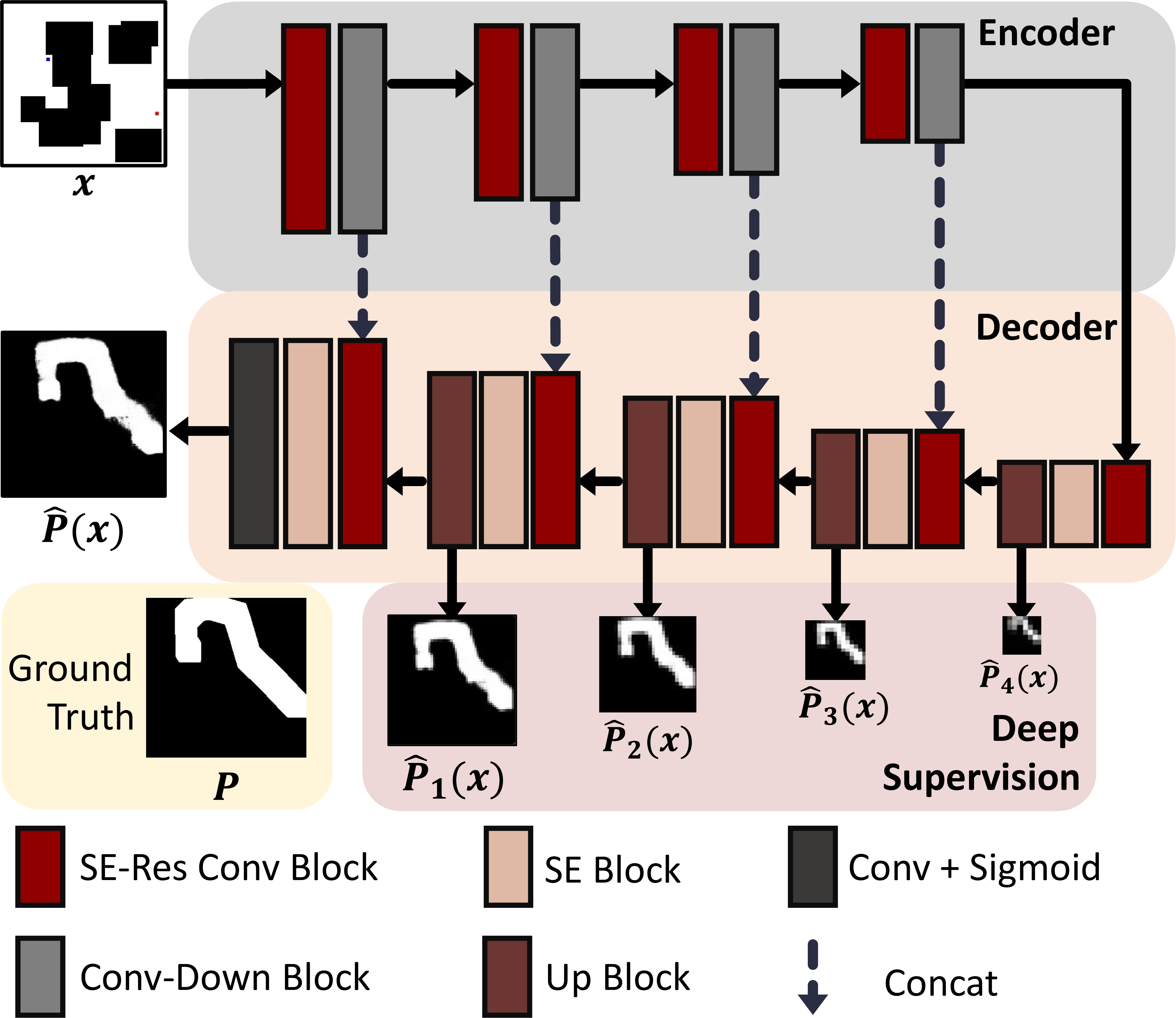}
	\caption{Structure of the RPNN, decomposed into an encoder, a decoder, and a deep supervision module at each hidden layer in the decoder.}
	\label{fig2}
	\vspace{-5.5mm}
\end{figure}

\subsection{Region prediction neural network}
As shown in Fig. \ref{fig2}, the RPNN is fed up with a 2-D RGB image as $x$, which represents an original map with a start vertex (red) and a goal vertex (blue). Specifically, the feasible configuration space is denoted in white, while the obstacle space is black. The output $\hat{P}(x)$ is a 2-D probability map, which has the same size as the input. Each pixel in the probability map with a value $p\in[0,1]$ reflects the likelihood of lying in the region. At hidden layers, the RPNN produces the side outputs denoted as $\widehat{P_{l}}:l\in[1,2,3,4]$, involving the comparisons with local outputs and the ground truth.
\begin{itemize}
\item \textit{Encoder:} There are four layers in the encoder, and each layer is comprised of a ``SE-Res Conv + Conv-down" module. As exhibited in Fig. 3, the SE-Res Conv block extracts a wealth of information with different levels via a residual convolution (denoted as Res Conv) block \cite{he2016deep} and a Squeeze-and-Excitation (SE) \cite{hu2018squeeze} block. The contextual information will be transmitted to the decoder through a concatenation operation. The SE-Res Conv block is first introduced in \cite{hu2018squeeze} (as shown in Fig. \ref{fig3}), which performs to adaptively evaluate and recalibrate the significance of channel-wise features by allocating weights. In Fig. \ref{fig3}, the convolution block (denoted as Conv block), as a form of \textit{Conv-BN-LeakReLU-Conv-BN}, is utilized to extract feature maps, while the \textit{LeakyReLU} represents a leaky rectified linear unit. \textit{Conv} represents a convolution operation, and \textit{BN} represents the batch normalization. To formulate the block, we take $x_i$ and $y_i$ as the input and output of the $i_{th}$ layer. Then, the output can be denoted by
\begin{equation}
	y_{i}=\sigma({\mathcal{R}}({\mathcal{F}}(x_{i},{W_{i}}))+W_{s} x_{i},
\end{equation}
where ${\mathcal{F}}(x_i,{W_i})$ represents the mapping in the convolutional operation, and $\mathcal{R}(\cdot)$ denotes the recalibration and embedding of the channel weights. $\sigma(\cdot)$ is the leaky ReLU operator. To unify the channels and realize an element-wise addition between the input and output, a linear projection is constructed through a 1$\times$1 convolution (denoted as $W_s$) to match the shape. In the Conv-down block, a convolutional layer is implemented with a stride equal to 2 to shrink the size of feature maps instead of a max pooling block. The shape of the input is denoted as $(H,W,C_N)$, and $C_N$ means $N$ channels. The structure of the encoder can be expressed as: $(H,W,C_{3} )-(H/2,W/2,C_{64} )-(H/4,W/4,C_{128} )-(H/8,W/8,C_{256} )-(H/16,W/16,C_{512} )$.
\begin{figure}[t!]
	\centering
	\includegraphics[width=3  in]{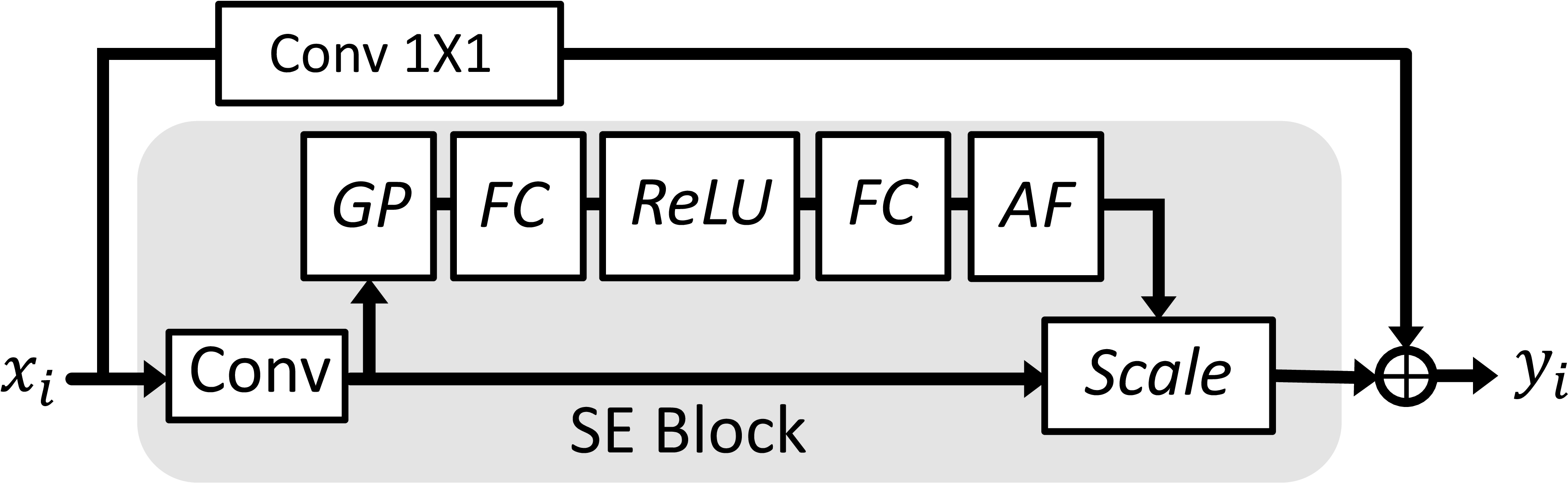}
	\caption{Illustration of the SE-Res Conv block. In the SE block, a global pooling operator (\textit{GP}) and a full connection layer (\textit{FC}) are equipped to compute the significance of each channel. \textit{AF} represents a Sigmoid-activated function. A form \textit{GP-FC-ReLU-FC-AF} performs to capture channel-wise significance fully. A scale operator multiplies the scalar of channel-wise relevance to the output of the convolution block before a short connection.}
	\label{fig3}
\end{figure}
\begin{figure}[t!]
	\centering
	\includegraphics[width=3  in]{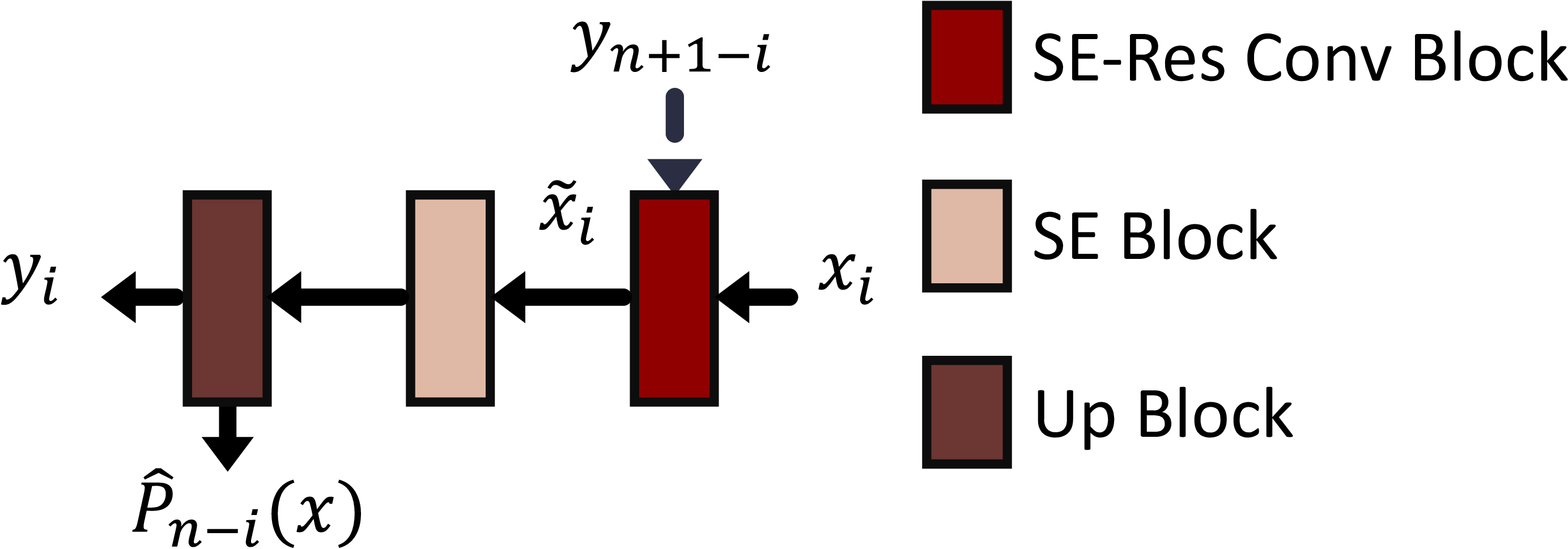}
	\caption{Illustration of the ``SE-Res Conv + SE + Up" module.}
	\label{fig4}
\end{figure}

\item \textit{Decoder:} There are five layers with a full-layer channel-wise attention module form of ``SE-Res Conv + SE + Up". As shown in Fig. \ref{fig4}, the diverse information $y_{(n+1-i)}$ is delivered from the encoder, and integrated into the input $x_i$ before the SE-Res Conv block. $n$ is denoted as the total number of layers in the model. The output of the SE-Res Conv block can be represented as
\begin{equation}
	\tilde{x}_{i}=\sigma\left({\mathcal{R}}\left({\mathcal{F}}\left(\left[x_{i}, y_{n+1-i}\right],\left\{W_{i}\right\}\right)\right)+W_{s}\left[x, y_{n+1-i}\right]\right.
\end{equation}
where $[\cdot]$ represents the fusion of multi-level feature maps. To disambiguate the negative effect of semantic gaps by the short connection, an additional SE block is implemented to reconstruct the relationship between channels after the SE-Res Conv block. The full-layer channel-wise attention filters out the polluted features with large semantic gaps, while the relevant features are maintained with large weights. Finally, the output of the ``SE-Res Conv + SE + Up" module can be defined as
\begin{equation}
	{y}_{i}={\mathcal{U}}({\mathcal{R}}(\hat{x}_i)),
\end{equation}
while $\mathcal{U}(\cdot)$ represents the up-sampling operator. Specifically, the final output of the decoder is produced by a convolutional operation and a sigmoid activation to obtain the predicted region. The structure of the decoder can be expressed as: $(H/16,W/16,C_{512})-(H/8,W/8,C_{512} )-(H/4,W/4,C_{256} )-(H/2,W/2,C_{128} )-(H,W,C_{64} )-(H,W,C_1 )$.

\item \textit{Multi-scale deep supervision:} To improve the transparency and robustness in the middle layers, a multi-scale deep supervision module is implemented. As depicted in Fig. \ref{fig4}, a side output  $\widehat{P_{l}}:l\in[1,2,3,4]$ is derived from the ``SE-Res Conv + SE + Up" module via a 1$\times$1 convolution operation and an activation function. Subsequently, the side outputs are resized and supervised, referring to the ground truth in the following procedure. The shape of output  $\widehat{P}_{l}(x)$ can be expressed as $(H*2^{-l},W*2^{-l},C_1)$.
\begin{figure*}[t!]
	\centering
	\includegraphics[width=6  in]{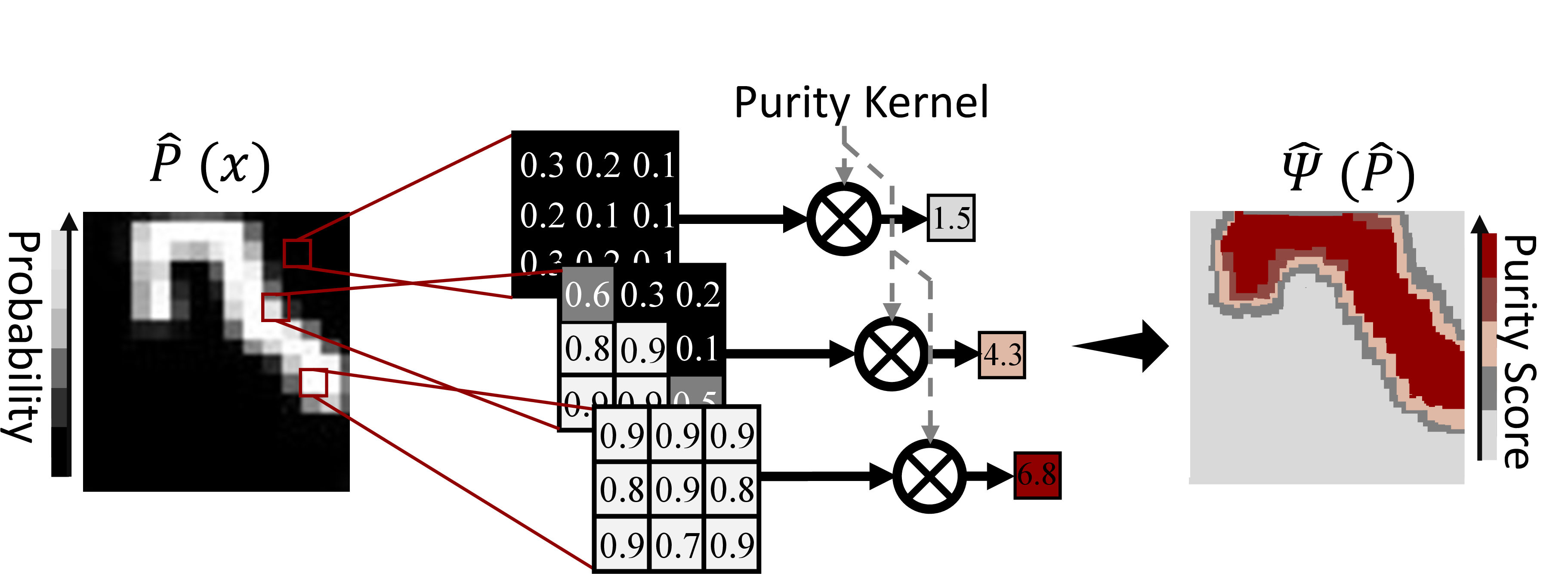}
	\caption{Demonstration of the estimated purity matrix.}
	\label{fig5}
	\vspace{1.5mm}
\end{figure*}
\end{itemize}
\subsection{Hybrid loss function}
The region prediction can be regarded as a semantic segmentation task. As a widely used loss function, binary cross entropy (BCE) performs to compare the pixel-level difference between the prediction result and the ground truth. Considering the fact that the central pixels in the ground truth count more for accuracy than the pixels on the edge, we devise a weighted BCE via a concept of purity. The purity is defined as a score to estimate the total count of positive or negative results in a pixel’s eight-neighbor. The purity score is computed by a convolution operation with a purity kernel $k_p$, while $k_p=\begin{bmatrix}
	1&1&1\\
	1&0&1\\
	1&1&1\\
\end{bmatrix}$. We deploy the convolution operation on the ground truth (denoted as $P(x)$) and the prediction result (denoted as $\hat{P}(x)$), respectively. The results are obtained as a purity matrix (denoted as $\mit\Psi(P)$) and an estimated purity matrix (denoted as $\mit\hat{\Psi}(\hat{P})$), which are in the same shape as the input. Significantly, the true positive pixels surrounded by the more positive pixels account more for the accuracy of the prediction results, while they reach higher scores in the estimated purity matrix, as shown in Fig. \ref{fig5}. Consequently, we assign a purity difference as weights to focus on the “hard-to-classify examples”. The pixel-level weighted BCE loss function ${\mathcal{L}}_{wBCE}$ can be summarized below:
\begin{equation}
	\begin{aligned}
		{\mathcal{L}}_{wBCE}=-\sum_{i=1}^{H}\sum_{j=1}^{W}w_{i,j}P_{i,j}\log({\hat{P}(x)}_{i,j})\\
		+w_{i,j}(1-P_{i,j})\log(1-{\hat{P}(x)}_{i,j}),
	\end{aligned}
\end{equation}
while the purity difference is denoted as 
\begin{equation}
	w_{i,j}=\|\mit\Psi_{i,j}-\mit\hat{\Psi}_{i,j}\|+\sigma_{smoothing}.
\end{equation}
$\sigma_{smoothing}$ is a constant value to avoid the difference being 0. A Dice loss function ${\mathcal{L}}_{Dice}$  is also employed to compare the relationship between the prediction result $\hat{P}$ and ground truth $P$ in map-level, defined by Eq. (6).
\begin{equation}
{\mathcal{L}}_{Dice}=1-\frac{2\|P\cap\hat{P}\|}{\|P\|+\|\hat{P}\|}
\end{equation}
To further emphasize the significant pixels, a purity loss ${\mathcal{L}}_{p}$ is defined to compare the pitch-level difference between the purity matrix and the estimated purity matrix. Then, the loss can be defined as
\begin{equation}
	{\mathcal{L}}_{p}=\sum_{i=1}^{H}\sum_{j=1}^{W}\|\mit\Psi_{i,j}-\hat{\mit\Psi}_{i,j}\|.
\end{equation}
Benefitting from the deep supervision module, a supervised loss based on the ground truth and side outputs can be formulated as
\begin{equation}
	{\mathcal{L}}_{sup}=\sum_{l=1}^{(n-1)/2}\beta_l*{\mathcal{L}}_{sup}(\mit\Pi^l(P),\widehat{P}_l),
\end{equation}
where $n=9$ is the number of total layers, $\beta_l=2^{(-(l-1))}:l\in[1,2,3,4]$ is a coefficient to balance the significance among different layers. $\text{$\mit\Pi$}^l(\cdot):\{0,1\}^{256\times256\times1}\rightarrow\{0,1\}^{2^{8-l}\times2^{8-l}\times1}$ is a downsampling operation to lessen the size of the ground truth.
Finally, the hybrid loss function is given by
\begin{equation}
	{\mathcal{L}}={\mathcal{L}}_{wBCE}+{\mathcal{L}}_{Dice}+\alpha*{\mathcal{L}}_{p}+{\mathcal{L}}_{sup},
\end{equation}
where $\alpha$ is a coefficient.

\subsection{RPNN-RRT*}
Same as other neural-network-driven methods, RPNN-RRT* utilizes the prediction results to realize a non-uniform sampling in the search procedure. A biased heuristic sampler is adopted with a probability $b_h$ (as shown in Algorithm 1). The uniform sampler guarantees the probability completeness of the RPNN-RRT*.

\section{Simulation Experiments}
In this section, we introduce the dataset CEMP and training details. Then, ablation studies on the structures and the loss functions are carried out. Besides, the RPNN is compared with other region prediction models. The region-wise accuracy performance is evaluated by the Dice coefficient, which can be expressed by the subtrahend in Eq. (6). Finally, the RPNN-RRT* is compared with other neural-network-driven methods in respect of the calculation time, the path cost, and the iteration number.
\subsection{Dataset CEMP and training details}
We build a dataset, Complex Environment Motion Planning (CEMP), with 16000 RGB examples with a size of 256$\times$256, including different scenarios with narrow passages. The ground truth is denoted by a concrete region, as the white regions are shown in $x$ in Fig. \ref{fig2}. The region is dilated from the optimal solution, which is computed by the RRT*. Instead of perforated regions with fuzzy boundaries, the dilated promising regions allow the model to learn edge information for accurately covering the optimal solutions. In addition, the typical evaluation metrics are available to compare the prediction results with the dilated ground truth.

All models are implemented with PyTorch on a 16 GB NVIDIA GeForce RTX3080 Ti Laptop GPU. We adopt an Adam optimizer with cosine annealing warm restarts. The dataset is split into a training dataset, a validated dataset, and a test dataset with a ratio of 8:1:1. Finally, the test dataset is simulated to demonstrate outstanding performance on accuracy and path planning.
\begin{table*}[!t]
	\caption{Ablation Study Results of the structure of the RPNN\label{tab:table1}}
	\vspace{-2.5mm}
	\centering
	\resizebox{\linewidth}{!}{
		\begin{tabular}{cccccccc}
			\hline
			\makecell[c]{UNet \\ (baseline)} &max pooling&Conv-down & SE5 & Conv  & Res Conv & SE-Res Conv  & \makecell[c]{Dice coefficient{(\%)}} \\
			\hline
			\checkmark&\checkmark&& &\checkmark& & &$\text{76.14}_{\pm\text{0.08}}$\\
			\checkmark& &\checkmark& &\checkmark& & &$\text{77.87}_{\pm\text{0.11}}$\\
			\checkmark& &\checkmark&\checkmark&\checkmark& & &$\text{87.76}_{\pm\text{0.14}}$\\
			\checkmark& &\checkmark&\checkmark& &\checkmark& &$\text{88.14}_{\pm\text{0.14}}$\\
			\checkmark& &\checkmark&\checkmark& & &\checkmark&$\textbf{88.24}_{\pm\textbf{0.05}}$\\
			\hline
	\end{tabular}}
	
\end{table*}
\textbf{\begin{table}[t]
		\caption{Ablation study results of the loss function\\ on the test dataset\label{tab:table2}}
			\vspace{-2.5mm}
		\centering
		\resizebox{0.8\linewidth}{!}{
			\begin{tabular}{cc}
				\hline
				Loss function & Dice coefficient(\%)\\
				\hline
				${\mathcal{L}}_{BCE}+{\mathcal{L}}_{Dice}$ & $\text{88.24}_{\pm\text{0.11}}$\\
				${\mathcal{L}}_{BCE}+{\mathcal{L}}_{Dice}+{\mathcal{L}}_{p}$ & $\text{86.50}_{\pm\text{0.13}}$\\
				${\mathcal{L}}_{wBCE}+{\mathcal{L}}_{Dice}$ & $\text{88.58}_{\pm\text{0.11}}$\\
				${\mathcal{L}}_{wBCE}+{\mathcal{L}}_{Dice}+{\mathcal{L}}_{p}$ & $\textbf{88.90}_{\pm\textbf{0.06}}$\\
				\hline
		\end{tabular}}
				\vspace{1.5mm}
\end{table}}
\textbf{\begin{table}[t]
		\caption{Region prediction results based on the test dataset\label{tab:table3}}
			\vspace{-2.5mm}
		\centering
		\resizebox{0.8\linewidth}{!}{
			\begin{tabular}{cc}
				\hline
				Region Prediction Model & Dice coefficient(\%)\\
				\hline
				$\text{UNet}$ & $\text{76.14}_{\pm\text{0.08}}$\\
				$\text{NEED}$ & $\text{83.29}_{\pm\text{0.13}}$\\
				$\text{MPT}$ & $\text{85.68}_{\pm\text{0.05}}$\\
				$\text{RPNN}$ & $\textbf{88.90}_{\pm\textbf{0.15}}$\\
				\hline
		\end{tabular}}
				\vspace{-5.5mm}
\end{table}}

\subsection{Comparison of region prediction}
Ablation study results of the RPNN model are listed in TABLE \ref{tab:table1}. All models use the BCE and Dice loss as the loss function. The models are embedded with different blocks, including a max pooling block, a Conv-down block, an SE5 block (indicating 5 SE blocks are added to the model), a Conv block, a Res Conv block, and a SE-Res Conv block. Hereafter the Dice coefficient is in a `mean$\pm$std' format. The result of the RPNN is highlighted in bold, which outperforms the other models. This result supports the notion that the collocation of the module ``SE-Res Conv + SE + Up" can achieve better performance by mitigating the harm of the semantic gaps. It should be noted that the improvement by the SE5 block is the most remarkable, indicating that the SE5 block alleviates the negative effect of the semantic gaps from the short connection.

Ablation study results of the proposed loss function are enumerated in TABLE \ref{tab:table2} based on the RPNN. The hybrid loss function of ``	${\mathcal{L}}_{wBCE}+{\mathcal{L}}_{Dice}+{\mathcal{L}}_{p}$" gains the highest Dice coefficient, which is boldfaced. It demonstrates the efficiency of the combination of the multi-level losses for assisting the model in learning the hierarchy features. Furthermore, the combination of ``${\mathcal{L}}_{BCE}+{\mathcal{L}}_{Dice}+{\mathcal{L}}_{p}$" perform worse than ``${\mathcal{L}}_{BCE}+{\mathcal{L}}_{Dice}$". The reason behind this observation is probably the model fails to involve joint learning of the multi-hierarchy losses.

\begin{figure}[t]
	\subfloat[Scenarios 1 to 5 from left to right.]{\includegraphics[width=3.3in]{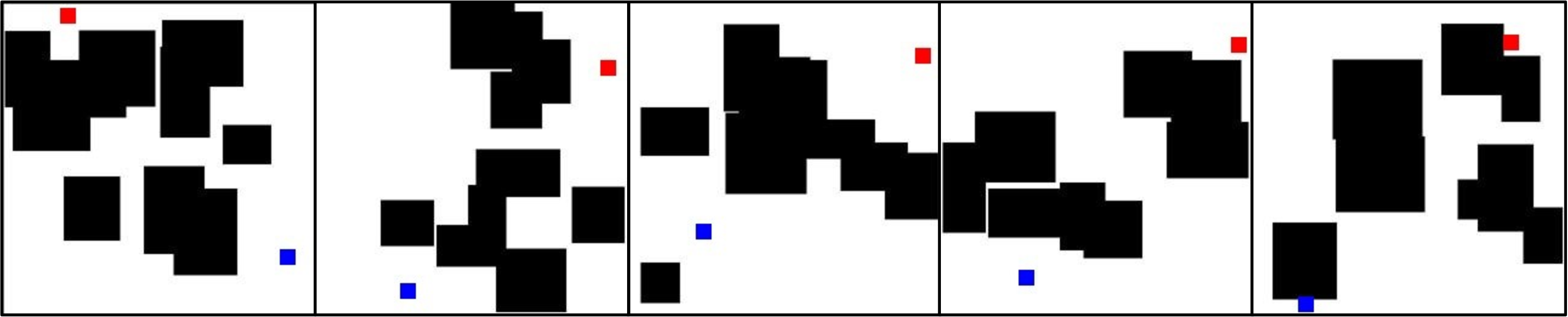}\label{fig6a}}
		\vspace{-0.1mm} 
	\subfloat[NEED prediction results.]{\includegraphics[width=3.3in]{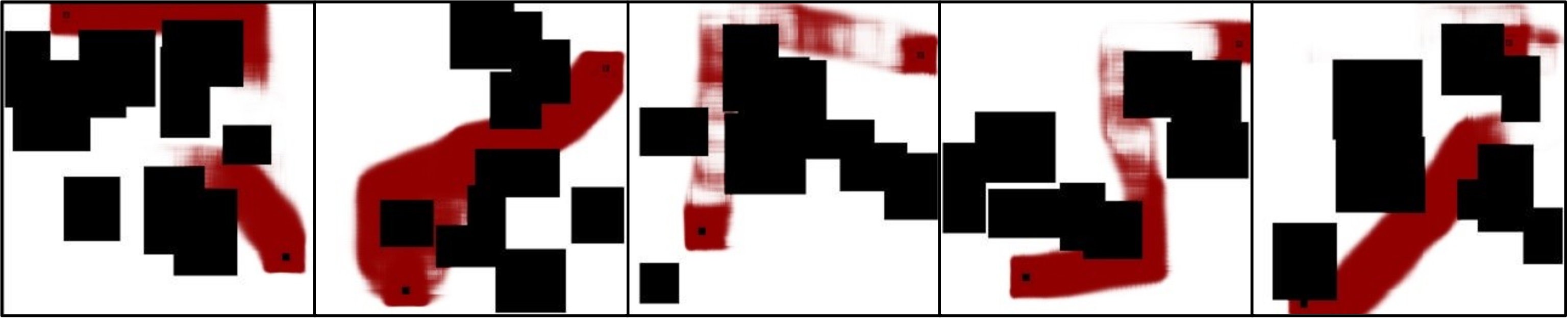}\label{fig6b}}
	\vspace{-0.1mm} 
	\subfloat[MPT prediction results.]{\includegraphics[width=3.3in]{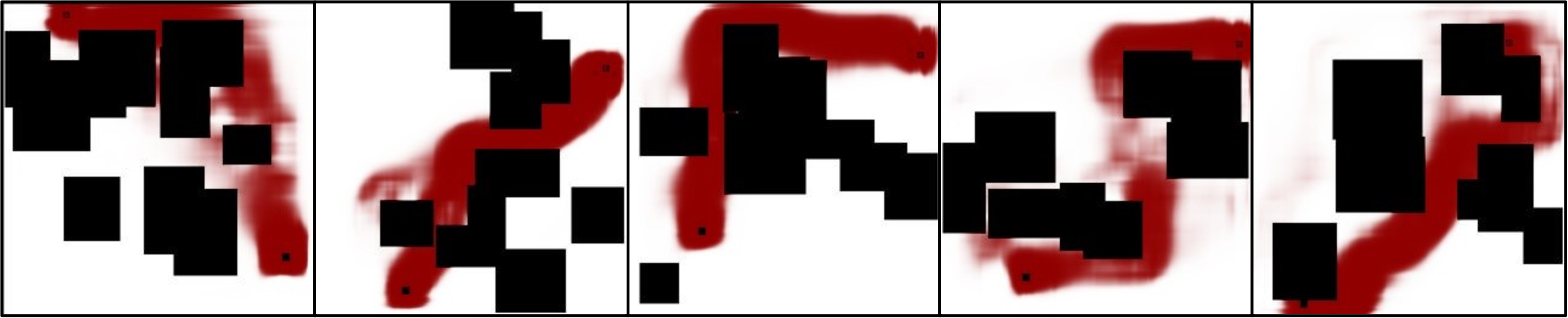}\label{fig6c}}
		\vspace{-0.1mm} 
	\subfloat[RPNN prediction results.]{\includegraphics[width=3.3in]{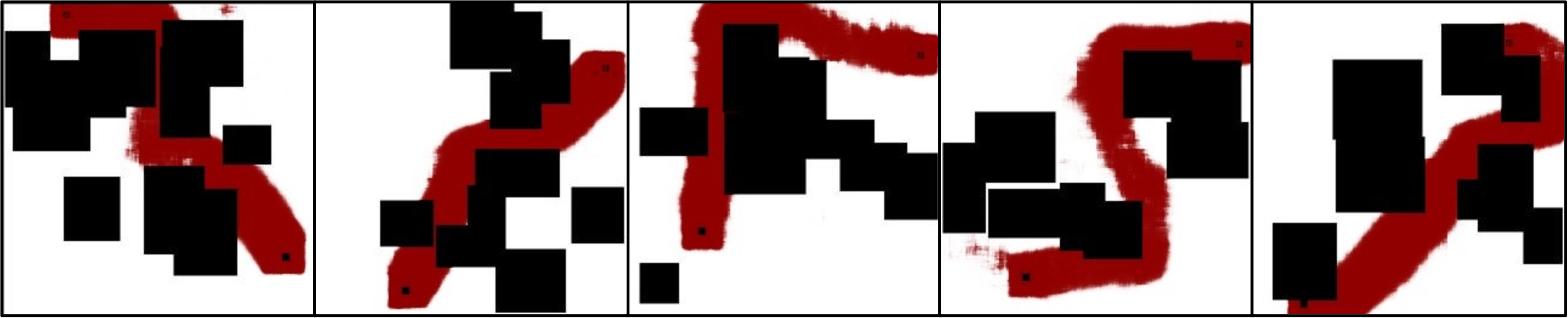}\label{fig6d}}
	\caption{Illustration of region prediction experiments based on the test dataset of the CEMP. Deep Red: predicted promising region; Black: obstacles.} 
			\vspace{-1.5em} 
\end{figure} 
We compare the RPNN with state-of-the-art region prediction models, Neural-network-driven Prediction (NEED) \cite{wang2021robot} and Motion Planning Transformers (MPT)\cite{johnson2021motion}. As presented in TABLE \ref{tab:table3}, the RPNN attains the highest Dice coefficient, which shows the improvement of the deep supervision module and the modifications of the model. Some scenarios are exhibited in Fig. 6. In all scenarios, the RPNN returns connected regions that accurately contain the optimal paths. In scenario 1, the NEED and the MPT even ignore the narrow passage and return disconnected regions. Low-quality and misleading regions fail to contain the optimal paths, which is lethal for non-uniform sampling. Interestingly, MPT is superior to NEED in most scenarios. The reason behind the observation is that MPT adopts a transformer structure instead of the encoder-decoder architecture, which avoids the semantic gap issue to some extent.

\subsection{Comparison of optimal path planning}

\begin{table*}[t!]\tiny
	\caption{Comparison results of RRT*, NEED-RRT*, MPT-RRT*, and RPNN-RRT* for optimal path planning\label{tab:table4}}
	\vspace{-1.5em}
	\centering
	\resizebox{0.8\linewidth}{!}{
		\begin{tabular}{cccccc}
			\Xhline{0.2pt}
			& & RRT* & NEED-RRT*&MPT-RRT*&RPNN-RRT*\\
			\Xhline{0.2pt}
			\multirow{3}*{Scenario 1}&Vertices&\text{1017}&\text{727}&\text{447}&\textbf{261}\\
			&Time (sec)&\text{0.93}&\text{1.07}&\text{0.33}&\textbf{0.09}\\
			&Success Rate&\text{98\%}&\text{28\%}&\text{74\%}&\textbf{100\%}\\
			\Xhline{0.2pt}
			\multirow{3}*{Scenario 2}&Vertices&\text{1712}&\text{1026}&\text{548}&\textbf{366}\\
			&Time (sec)&\text{2.29}&\text{1.26}&\text{0.50}&\textbf{0.23}\\
			&Success Rate&\text{82\%}&\text{98\%}&\text{98\%}&\textbf{100\%}\\
			\Xhline{0.2pt}
			\multirow{3}*{Scenario 3}&Vertices&\text{1308}&\text{1314}&\textbf{404}&\text{522}\\
			&Time (sec)&\text{1.53}&\text{2.00}&\text{0.16}&\textbf{0.11}\\
			&Success Rate&\textbf{100\%}&\text{84\%}&\textbf{100\%}&\textbf{100\%}\\
			\Xhline{0.2pt}
			\multirow{3}*{Scenario 4}&Vertices&\text{1620}&\text{904}&\text{595}&\textbf{448}\\
			&Time (sec)&\text{2.13}&\text{1.02}&\text{0.46}&\textbf{0.20}\\
			&Success Rate&\text{90\%}&\text{82\%}&\textbf{100\%}&\textbf{100\%}\\
			\Xhline{0.2pt}
			\multirow{3}*{Scenario 5}&Vertices&\text{1934}&\text{565}&\text{575}&\textbf{510}\\
			&Time (sec)&\text{2.63}&\text{0.44}&\text{0.38}&\textbf{0.27}\\
			&Success Rate&\text{70\%}&\text{84\%}&\text{98\%}&\textbf{100\%}\\
			\Xhline{0.2pt}
			
	\end{tabular}}
\end{table*}

To assess the efficiency of the RPNN-RRT* algorithm in optimal path planning, we conduct experiments using the examples in the test dataset. As a comparison, neural-network-driven methods, NEED-RRT* and MPT-RRT*, are performed based on the respective prediction results. Each experiment is run fifty times independently. The algorithm is terminated if the cost of the generated path differed by 3\% or less from the length of the optimal path within 5000 samples, at that point the current solution is returned. 

\begin{figure}[t]
	\centering
	\subfloat[RRT*.]{\includegraphics[width=3.5in]{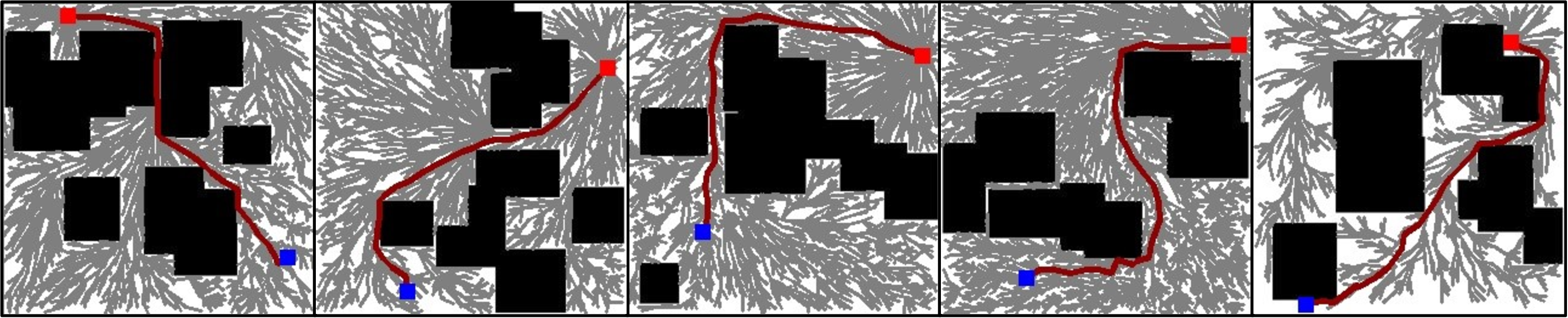}\label{fig7a}}
	\vspace{-0.1mm} 
	\subfloat[NEED-RRT*.]{\includegraphics[width=3.5in]{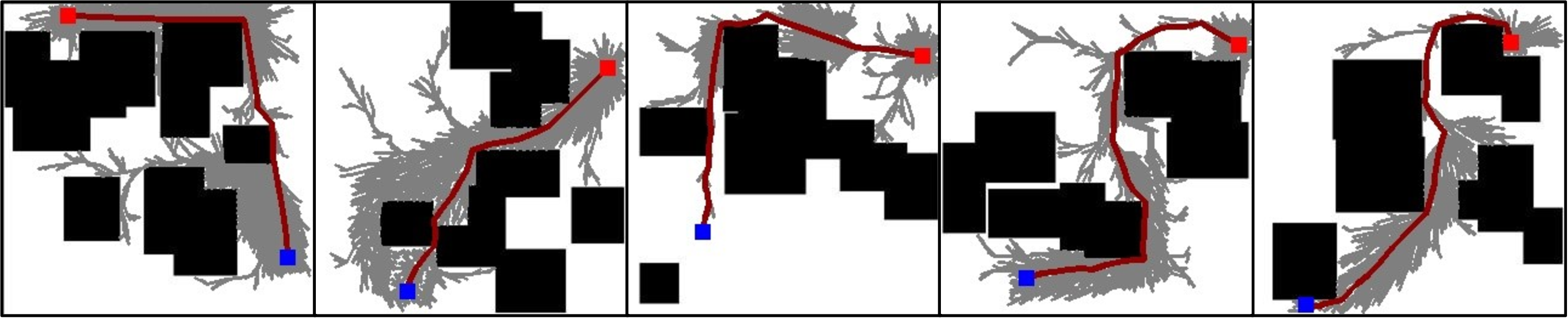}\label{fig7b}}
	\vspace{-0.1mm} 
	\subfloat[MPT-RRT*.]{\includegraphics[width=3.5in]{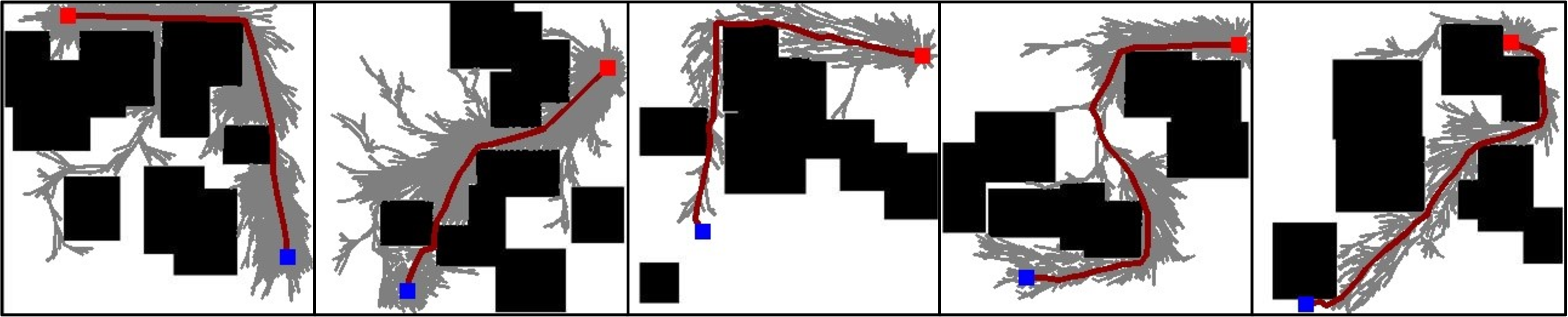}\label{fig7c}}
	\vspace{-0.1mm} 
	\subfloat[RPNN-RRT*.]{\includegraphics[width=3.5in]{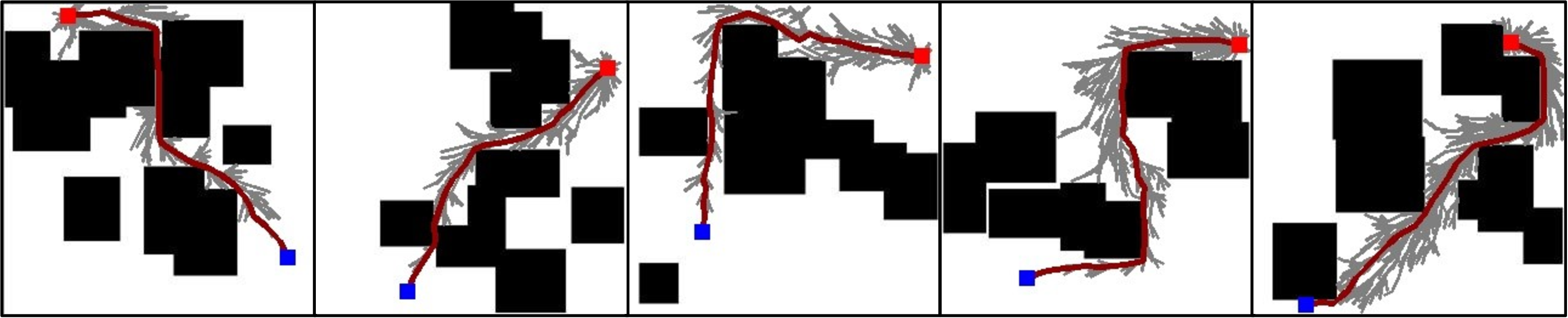}\label{fig7d}}
	\caption{Illustration of the vertices distributions and exploration areas. Vertices are at the end of the gray branches. Deep red line represents the final path.} 
	\vspace{-1.5em} 
\end{figure} 
\begin{figure}[t]
	\centering
	\includegraphics[width=3.3  in]{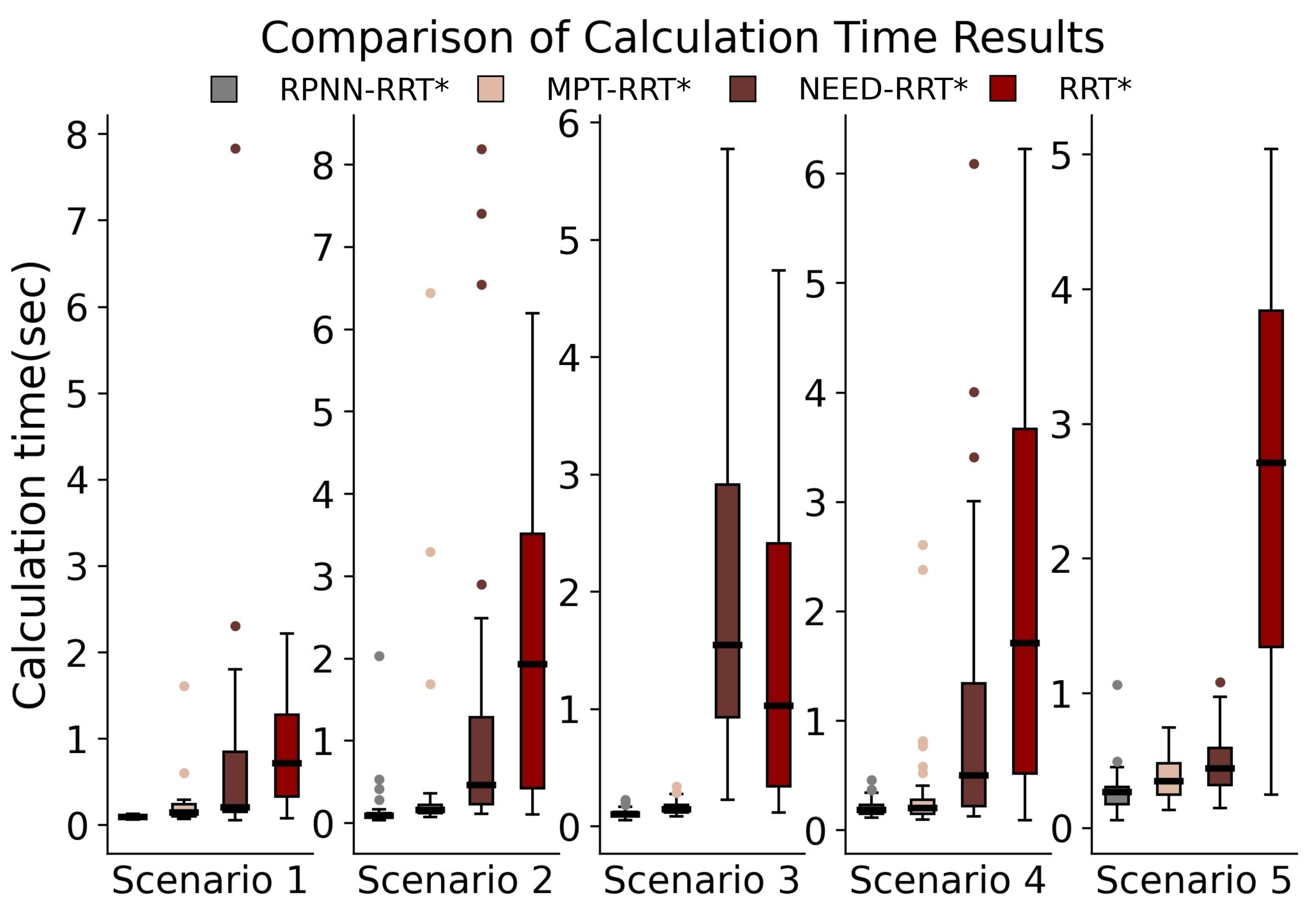}
	\caption{Calculation time comparison of the RPNN-RRT* and the other methods. The bold black line in the box indicates the median calculation time, while the error bar presents the maximum and minimum calculation time within 1.5 IQRs (interquartile range) of the first and third quartile. The round dots are outliers, which represent occasional cases.}
	\label{fig8}
\end{figure}

In addition to the average calculation time, Table \ref{tab:table4} exhibits the number of vertices and the success rate. The best results are highlighted in bold. The results show that the RPNN-RRT* can find the optimal path with fewer vertices and fewer attempts than the other neural-network-driven methods. The observation demonstrates that the promising regions perform as guidance and efficiently generate valuable vertices for exploration trees by narrowing the sampling domain. Besides, the RPNN-RRT* achieves a great reduction in calculation time by 29-73\% and 90-91\%, respectively, compared with the other neural-network-driven methods and the RRT*. It should be noted that the success rate is closely associated with the quality of the regions, which accounts for the low success rate of the NEED. 

Fig. 7 depicts the vertices distributions and exploration areas. Due to the low-accuracy regions, the NEED-RRT* and the MPT-RRT* attempt to search in a wider area, especially in scenario 1. In some cases, uniform sampling assists the exploration to overcome the influence of the disconnected regions, although the path deviates from the optimal path. To further demonstrate the efficiency of the RPNN-RRT*, we compare the median calculation times of the RNPP-RRT*, the MPT-RRT*, the NEED-RRT*, and the RRT* in 50 experiments. The results, shown in Fig. \ref{fig8}, indicate that the RPNN-RRT* consistently outperforms the other methods in calculation time in all scenarios. In contrast, the RRT* exhibits poorer performance due to its uniform sampling strategy. The NEED-RRT* struggles to find the optimal path, which is caused by the inaccurate regions as the heuristics. Moreover, the MPT-RRT* exhibits close performance on the median calculation time.

\section{Conclusion}
In this work, we focused on improving the accuracy of the region prediction for neural-network-driven optimal path planning algorithms, further enhancing the sampling efficiency and reducing the calculation time. We designed an encoder-decoder model, Region Prediction Neural Network (RPNN), to predict high-accuracy regions. Channel-wise attention module was equipped to narrow the semantic gap in the short connection. Furthermore, a hybrid loss based on purity was devised to assist the learning to concentrate on more significant pixels. Finally, we proposed a neural-network-driven method for optimal path planning, abbreviated to RPNN-RRT*. Simulation results demonstrated that 89.13\% accuracy was achieved in region prediction based on a Complex Environment Motion Planning (CEMP) dataset. The calculation time for the optimal path planning was markedly reduced by 29-73\% compared with the other neural-network-driven methods with fewer vertices and a higher success rate.

The combination of neural networks and path planning is an interesting topic for future work. It could continue to explore the accuracy improvement, extend to other sampling-based path planning algorithms, and apply it to more practical scenarios. 


\end{document}